\documentclass{article}
\pdfoutput=1

\usepackage[final]{neurips_2023}
\usepackage[utf8]{inputenc} % allow utf-8 input
\usepackage[T1]{fontenc}    % use 8-bit T1 fonts
\usepackage{hyperref}       % hyperlinks
\usepackage{url}            % simple URL typesetting
\usepackage{booktabs}       % professional-quality tables
\usepackage{amsfonts}       % blackboard math symbols
\usepackage{nicefrac}       % compact symbols for 1/2, etc.
\usepackage{microtype}      % microtypography
\usepackage{xcolor}         % colors
\usepackage{multicol}
\usepackage{algorithm,algpseudocode}
\usepackage{dsfont}
\usepackage{multirow}

\usepackage{graphicx}
\usepackage{amsmath}
\usepackage{bm}
\usepackage{amssymb}
\usepackage[normalem]{ulem}

\newcommand{\set}[1]{\mathcal{#1}}

\newtheorem{NThm}{Theorem}[section]

\newtheorem{NDef}[NThm]{Definition}

\newtheorem{NRem}[NThm]{Remark}

\title{Marked Neural Spatio-Temporal Point Process Involving a Dynamic Graph Neural Network\thanks{Updated version.}}

\author{%
  Alice Moallemy-Oureh\thanks{All contributions are listed in detail at the end of this paper.}, \; Silvia Beddar-Wiesing, \; Yannick Nagel, \; Rüdiger Nather, \; Josephine Thomas \\
    University of Kassel, 
  34121 Kassel, Germany \\
  \texttt{\{amoallemy, s.beddar-wiesing, yannick.nagel, r.nather, jthomas\}@uni-kassel.de } \\
}

\begin{document}

\maketitle
\vspace{-.2cm}
\begin{abstract}
  Temporal Point Processes (TPPs) have recently become increasingly interesting for learning dynamics in graph data. A reason for this is that learning on dynamic graph data is becoming more relevant, since 
data from many scientific fields, ranging from mathematics, biology, social sciences, and physics to computer science, is naturally related and inherently dynamic. In addition, TPPs provide a meaningful characterization of event streams and a prediction mechanism for future events. 
%While training Recurrent Neural Networks and solving PDEs for representing temporal data is expensive, Temporal Point Processes (TPPs) appeared to be suitable alternatives. 
%A TPP is characterized by a conditional density function over time that determines the intensity of event occurrences. The construction of the intensity is dependent on assumptions about particular temporal behavior in the data. 
Therefore, (semi-)parameterized Neural TPPs have been introduced whose characterization can be (partially) learned and, thus, enable the representation of more complex phenomena. However, the research on modeling dynamic graphs with TPPs is relatively young, and only a few models for node attribute changes or evolving edges have been proposed yet. To allow for learning on fully dynamic graph streams, i.e., graphs that can change in their structure (addition/deletion of nodes/edge) and in their node/edge attributes, we propose a Marked Neural Spatio-Temporal Point Process (MNSTPP). It leverages a Dynamic Graph Neural Network to learn a Marked TPP that handles attributes and spatial data to model and predict any event in a graph stream.

%It is necessary, however, to make assumptions of particular temporal behavior in the data for constructing an appropriate TPP for a given observed sequence. 
\end{abstract}

\section{Introduction}
Learning on graphs is part of state-of-the-art research in Machine learning. In this context, GNNs have emerged as the baseline models for integrating structural information into the learning process. Many models have been developed for static graph data that can use structural information in the learning process of various problems \cite{unser_survey}. However, many applications produce more complex data and problems, necessitating the incorporation of temporal information, which static graphs cannot capture \cite{book_hamilton_2020}. 
Therefore, learning on dynamic graphs has been growing for several years. Models have been proposed that can handle node attribute changes \cite{inproc_seo_2018, STGAT, DGCRN}, additions of nodes or edges (growing graphs) \cite{inproc_trivedi_2017,inproc_trivedi_2019, han2020graph}, edge-structure dynamics and node attribute changes \cite{arx_wang_2020, EasyDGL, arx_rossi_2020}, or node and edge additions and deletions in graph snapshot sequences \cite{najafi2023entropy}. More examples are provided in \cite{jour_kazemi_2020,unser_survey}. This, however, demonstrates that most models are specialized for processing certain dynamic graphs on specific graph representations.

Approaches to modeling dynamics in graphs can be divided into three main categories: The usage of Recurrent Neural Networks (RNNs), Partial Differential Equations (PDEs), or Temporal Point Processes (TPPs). However, the drawback of RNNs is that explicit temporal information is neglected, and only the sequence of changes in the graph is represented. Furthermore, solving PDEs is costly and non-trivial, making TPPs more appropriate for representing graph stream data. However, they have the disadvantage of making assumptions about the temporal patterns that are not necessarily met. Therefore, Neural TPPs have been proposed \cite{First_NTPP, NeuralHawkes}, which can efficiently learn the temporal evolution of complex processes without requiring prior knowledge. Though only a few models for graphs have been developed exploiting them yet, they are also limited to certain graph types.

In this work, we present the first Marked Neural Spatio-Temporal Point Process (MNSTPP) utilizing a Dynamic GNN to learn on graph streams with arbitrary structural and attribute dynamics. The approach is based on the model DyREP \cite{inproc_trivedi_2019}, which applies a Neural TPP on edge-growing graphs. We extend the Neural TPPs of DyREP to Marked Neural Spatio-Temporal Point Processes, which support processing any structural changes in graphs. The representation of continuous node and edge attributes in the form of a marked TPP with real-valued marks is innovative compared to the usage of non-marked TPPs or categorical marks as carried out in most of the literature so far \cite{chang2020continuous,xia2022graph} and \cite{gracious2023neural, NeuralHawkes}, or marks only for non-graph data \cite{STPP1, MultistreamMPPCumHazard, MultiStreamMPP, zhou2022neural}. By creating the Neural TPPs based on a Dynamic GNN, we can predict all kinds of structural events, event times, and attribute changes over time. Furthermore, the model is updateable for every new event by local retraining, which enables a fast and efficient update and training of the model.

% \textcolor{red}{The experiments are conducted on X datasets. We compare the X measure against X baseline models. Since there are only a few available TPP-based models, and none that can process fully dynamic graphs, we test our model against different baseline models for different problems. The results show X. Future work is X.}

\section{Foundations}
Before we go into more detail about the structure of our model and explain its individual modules, we provide an introduction to the essential concepts here. Additionally, we give an overview of the used notation.
\begin{table}[h!]
    \centering
    \caption{Notation and nomenclature.}
    \begin{tabular}{||c|c||}
        \hline\hline
        $\set{V}_t$ & set of nodes at time $t$\\ \hline
        $\set{E}_t$ & set of edges at time $t$\\ \hline
        $\alpha_t,\,\beta_t$ & attribute mappings at time $t$\\ \hline
        $\set{T}$ & time interval\\ \hline
        $\bar{t}$ & time of last event before $t$\\\hline
        $x\in\set{V}\cup\set{E}$ & item: node or edge \\\hline
        $k$ & event type $\{0,\ldots,5\}$\\\hline
        $\gamma_t(x)$ & attribute of item $x$ at time $t$ \\\hline
        $\bm{H}_{\bar{t}}$ & event history up to time $\bar{t}$\\\hline
        $\set{N}(x)$ & neighborhood of $x$\footnotemark\\ \hline
         $[\,\cdot\,|\,\cdot\,]$ & vertical concatenation \\\hline
         $d_h$ & hidden dimension \\\hline
         $d_{\gamma}$ & attribute dimension of $\gamma$\\\hline
         $\sigma$ & sigmoid activation function \\
         \hline\hline
    \end{tabular}
    \begin{tabular}{||c|c||}
        \hline\hline
        $\|\cdot\|$ & vector norm \\\hline
        $\lambda(t,x,k,\gamma)$ & intensity of event $(t,x,k,\gamma)$\\\hline
        $g(\cdot)$ & scoring function \\\hline
        $[a],\,[a]_0$ & $\{1,\ldots,a\},\,\{0,\ldots,a\}$ \\\hline
        $\phi(\cdot)$ & modified softplus function\\\hline
        $\psi_k$ & time scaling of event type $k$\\\hline
        $\bm{M}_*,\,\bm{m}_*$ & DGNN model parameter\\\hline
        $q_*$ & attention factor\\\hline
        $\bm{W}_*,\,\bm{w}_*$ & scoring parameters\\\hline
        $\bm{Z}_t^x$ & embedding of item $x$ at time $t$\\\hline
        $a|_b$ & $a$ restricted to $b$\\\hline
        $\set{O},\,\tilde{\set{O}}$ & observations and negative samples\\\hline
        $f,\, F$ & conditional density and the CDF\\
         \hline\hline
    \end{tabular}
    \label{tab:notation}
\end{table}

\footnotetext[2]{The neighborhood of a node contains the adjacent nodes, while the neighborhood of an edge includes the incident nodes.}
\subsection{Graph Streams}

A \textbf{graph stream} $G = \left(\mathcal{G}_0, \mathcal{O}\right)$ is a dynamic graph in continuous-time representation defined by a static start graph $\mathcal{G}_0$ and an event stream $\mathcal{O}$. The start graph $\mathcal{G}_0=(\mathcal{V}_0, \mathcal{E}_0, \alpha_0, \beta_0)$ contains a finite node set $\mathcal{V}_0\subset\mathbb{N}$, (undirected) edges  $\mathcal{E}_0=\{\{u,v\}\mid u,v\in\set{V}_0\}$ between them, and attribute mappings ${\alpha_0: \mathcal{V}_0 \rightarrow \mathcal{A},} \; {\beta_0: \mathcal{E}_0 \rightarrow \mathcal{B}}$ with arbitrary attribute sets $\mathcal{A}, \mathcal{B}$ and, w.l.o.g., $\mathcal{A},\mathcal{B}\subseteq\mathcal{C}$. 
% \begin{figure}[h!]
%     \centering
%     \includegraphics[width=.5\linewidth]{figures/sdg.pdf}
%     \caption{\textcolor{red}{TBD: fully dynamic graph}}
%     \label{fig:graph_stream}
% \end{figure}

The observations in the event stream $\set{O}$ are of the form $(x,t,k,\gamma_t(x))$ and determine on which item $x$ (node or edge) an event of type $k$ (addition, deletion or attribute change) with attribute $\gamma_t(x)$ happens at time $t\in\set{T}\subset\mathbb{R}_0$. Here, we assume to have unique timestamps for each event, to enable a proper update\footnote{This is important because of the embedding updates in the model. Everytime a node embedding changes due to new events, it influences the neighborhood embeddings. Consequently, if there were simultaneous events, the update sequence would no longer be unique.} of the model. 
% As example, Fig.~\ref{fig:graph_stream} shows an event stream of node and edge additions and deletions together with attribute changes.

%\input{preliminaries_subsections/DGNN}
\subsection{Temporal Point Processes}
%A sequence of observations can be regarded as a random process. Each event at a point in time is then treated as a random variable, which in turn depends on the previous observations. This is referred to as a Temporal Point Process (TPP). Temporal Point Processes provide a conditional density function that acts as a function over time and comes with an approach to predict new events in the future. 
A sequence of observations forms a temporal point process (TPP), where each event at a given time point is treated as a random variable influenced by previous observations. TPPs offer a conditional density function that operates over time, enabling the prediction of future events.
%can be characterized as a counting function quantifying the number of events occuring at points in time. 
%The model proposed here is a specific TPP, a \textbf{Marked Neural Spatio-Temporal Point Process (MNSTPP)}, to utilize these advantages. It enables processing event streams including attributed and spatial events in a graph without prior knowledge about the temporal behavior of the stream. In the following, the Marked and Spatio-Temporal Point Process are defined for attributed (marked) and spatial events. Afterwards, the concept of Neural Temporal Point Processes is introduced that relieves the assumptions of specific temporal patterns in the process and enables the learning of the temporal evolution of an input stream.
The proposed model is a specialized type of temporal point process known as a \textbf{Marked Neural Spatio-Temporal Point Process (MNSTPP)}. It capitalizes on these benefits by handling event streams comprising attributed and spatial events within a graph without requiring prior knowledge about the temporal dynamics of the stream.
Firstly, the definitions of Marked and Spatio-Temporal Point Processes are presented for attributed and spatial events, respectively. Subsequently, the concept of Neural Temporal Point Processes is introduced, which relaxes assumptions about specific temporal patterns in the process and enables learning the temporal evolution of an input stream.

\begin{paragraph}{Marked Temporal Point Processes.}
Given an event stream, it is non-trivial to tell when a new event will occur and what it may look like. Given that the event is dependent on the historical point patterns, the underlying process can be modeled as a \textbf{Temporal Point Process (TPP)} \cite{TPP_lecture}. Furthermore, the events in a graph stream as defined above include the timestamps together with additional information as the item, the event type and an attribute. Consequently, the process is modeled as a \textbf{Marked TPP (MTPP)}, where the marks $\bm{m}$ in the following comprise the additional information, so $\bm{m} =(x,k,\gamma_t(x))$. As a result, the marks can be discrete, continuous or a combination of both. Here, the marks in the MTPP are assumed to be i.i.d., i.e., a mark is only dependent on the timestamp but independent from its surrounding in space.

According to Campbell's theorem \cite{STPP_book}, the intensity $\lambda$ of a marked event $(t,x,k,\gamma)$ can be determined by the expectation of a \textbf{counting function} of $\set{H}$ over the compact set $[t,t+\Delta t)\times\set{B}(\bm{m}, \Delta \bm{m})$. The counting function for MTPPs is defined as
\begin{align}
    %\lambda(t,x,k,\gamma) = \mathbb{E}\bigl[N\bigl([t,\Delta t),\set{B}(\bm{m})\bigr)\bigr] \text{ with}\\
    N([t,t+\Delta t),\set{B}(\bm{m}, \Delta \bm{m})) = \sum\limits_{i\in [n]}\mathds{1}_{[t,t+\Delta t)}(t_i)\cdot\mathds{1}_{\set{B}(\bm{m}, \Delta \bm{m})}(\bm{m}_i).
\end{align}

\begin{NDef}[Marked Temporal Point Process]
Let $\set{H}=\{(t_i, \bm{m}_i)\}_{i\in[n]_0}$ be a sequence of $n> 0$ events with marks $\bm{m}_i\in\mathcal{M}$ in space $\mathcal{M}$ and timestamps $t_i\geq 0$. Further, let $N(t, \bm{m})$ be the joint counting process that counts the number of events with mark $\bm{m}$ up to time $t$ and ${\set{H}_{\bar{t}}=\{(t_i, \bm{m}_i)\mid t_i\leq\bar{t},\ (t_i, \bm{m}_i)\in \set{H}\}}$ entail the history of events before timestamp $t$ up to time $\bar{t}<t$.

The intensity function characterizes the MTPP completely \cite{TPP_lecture} and describes the instantaneous rate of events with a specific mark occurring at time $t$ defined by:
\begin{equation}\label{eq_MTPP}
    \lambda(t, \bm{m}\mid\mathcal{H}_t) = \lim_{\Delta t \downarrow 0, \Delta \bm{m} \downarrow 0} \frac{\mathbb{E}\Bigl[N\bigl([t, t + \Delta t)\times \set{B}(\bm{m}, \Delta \bm{m})\bigr) \mid \mathcal{H}_t\Bigr]}{\Delta t \cdot \nu(\Delta \bm{m})}.
\end{equation}
Here, $\set{B}(\bm{m}, \Delta \bm{m})\subset\mathbb{R}^d$ is an open hypersphere with center $\bm{m}$ and radius $\Delta \bm{m}$ for continuous marks and $\set{B}(\bm{m})=\bm{m}$ for discrete marks. Note that, if $\bm{m}$ is a combination of discrete and continuous marks, then $\set{B}(\bm{m})$ and $\Delta \bm{m}$ are defined component-wise.  Further,  $\nu(\Delta m)$ is the volume measure in the mark space. 

%%%%%%%%
%Let $\set{H}=\{(t_i, \bm{m}_i)\}_{i\in[n]_0}$ be a sequence of $n> 0$ events with marks $\bm{m}_i\in\mathcal{M}$ in space $\mathcal{M}$ and timestamps $t_i\geq 0$. Further, let ${\set{H}_{\bar{t}}=\{(t_i, \bm{m}_i)\mid t_i\leq\bar{t},\ (t_i, \bm{m}_i)\in \set{H}\}}$ entail the history of events before timestamp $t$ up to time $\bar{t}<t$.
%Then, the \textbf{conditional intensity} for a new event $(t^*,\bm{m}^*)$ to fall into $[t, t+\Delta t)\times\set{B}(\bm{m})$ given the history is given by
%\begin{align}\label{eq_MTPP}
%    \lambda(t, \bm{m}\mid \set{H}_{\bar{t}}) = \mathbb{P}\Bigl(t^*\in [t, t+\Delta t), \bm{m}^*\in \set{B}(\bm{m})\mid \set{H}_{\bar{t}}\Bigr)\cdot\mathbb{P}\Bigl(t^*\notin (-\infty,t)\mid \set{H}_{\bar{t}}\Bigr)
%\end{align}
%and characterizes the MTPP completely \cite{TPP_lecture}. 

\end{NDef}
%The intensity $\lambda$ is determined by the probability of the event to fall into a given ball in the attribute space and time, together with the probability that the event did not happen before (independent whether the attribute already appeared or not). 
Since the intensity function represents the instantaneous event rate, reflecting the likelihood of an event to occur at a specific time with a specific mark given the historical context, the intensity function must remain non-negative.  
Fig.~\ref{fig:mtpp} illustrates the probability of an event to fall into a time and attribute interval.
\begin{figure}[ht!]
    \centering
    \includegraphics[width=.7\linewidth]{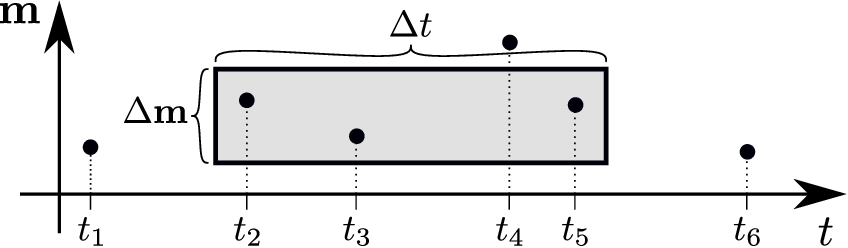}
    \caption{The figure is inspired by \cite{STPP_book}. The shaded rectangle is the ball for which we want to have the probability and the dots over time visualize the observed events.}
    \label{fig:mtpp}
\end{figure}

\begin{NRem}[Spatio-Temporal Point Process]
If the marks of the events $\bm{m}$ determine a location, such as a point in $\mathbb{R}^d$ or a node in a graph, the \textit{Spatio}-Temporal Point Process describes the evolution of data in space and time \cite{STPP}.
\end{NRem}
\end{paragraph}

\begin{paragraph}{Neural Temporal Point Processes.}
%The intensity function of a Point Process has to be chosen carefully based on assumptions about the appearance of temporal patterns in the data. Since these assumptions are not necessarily met in real-world data, partially and fully trainable Neural TPPs \cite{MultiStreamMPP} have been proposed \cite{First_NTPP, NeuralHawkes} to learn the more complex temporal pattern in the data. In the proposed work, we utilize a \textit{partially trainable Neural TPP}, so the definition of the fully-trainable Neural TPP is skipped here but can be found in \cite{MultiStreamMPP}.
The choice of the intensity function in point processes requires careful consideration, typically based on assumptions about the temporal patterns present in the data. However, real-world data often deviates from these assumptions. 
To address this challenge, partially and fully trainable (semi-)parameterized Neural TPPs have been introduced in the literature \cite{First_NTPP, MultiStreamMPP, NeuralHawkes}.
Here, we leverage a semi-parameterized Neural TPP. While the detailed definition of a fully parameterized Neural TPP is omitted here, it can be found in \cite{MultiStreamMPP}.

\begin{NDef}[Neural Temporal Point Process]
Semi-parameterized Neural TPPs \cite{MultiStreamMPP} replace intensity functions by Deep Neural Networks (DNNs). These encode the historical information $\set{H}_{t}$ of an MTPP in a feature vector $\set{H}_{t}$ which is then forwarded to a selected temporal decay function $\phi$. Together, they determine the intensity 
\begin{align}\label{eq_NTPP}
    \lambda(t, \bm{m}\mid \set{H}_{t}) & = \phi(\Delta t, \Delta \bm{m}\mid \set{H}_{t}) %\nonumber
     =\phi(\Delta t, \Delta \bm{m}\mid \text{\textnormal{DNN}}(\set{H}_{t})).
\end{align}
Here, $\Delta t = t - \bar{t}$ and $\Delta \bm{m} = \|\bm{m} - \bar{\bm{m}}\|$ are the deviations of the timestamps and marks from the events at $t$ and the last event before it.
\end{NDef}
\end{paragraph}

\section{Model}

Given a dynamic graph in form of a graph stream including structural as well as attribute changes, we aim to learn the underlying process of event occurrences on the graph over time to predict future events. For this purpose, we design a semi-parameterized \textbf{Marked Neural Spatio-Temporal Point Process (MNSTPP)} by integrating a Dynamic Graph Neural Network (DGNN) into the definition of a conditional intensity function. 

The idea of the MNSTPP is based on DyREP \cite{inproc_trivedi_2019}, a representation learning framework that characterizes communication and association dynamics between nodes in a graph separately with the aid of NTPPs. In our work, we provide one NTPP for each of the event types to learn their individual temporal behavior and to exploit the entire information given in a graph stream. 

The MNSTPP is approximated by the conditional intensity function
\begin{align}\label{eq:MNSTPP}
    \lambda\bigl(x, t, k, \gamma_t(x)\mid \bm{H}_t\bigr) = \sigma\Biggl(\phi\Bigl[g\left(x, t, k, \gamma_t(x)\mid \bm{H}_t\right), k\Bigr]\Biggr)
\end{align}
that maps an event $(x,t,k,\gamma_t(x))$ with item $x$, time $t$, event type $k$ and attribute $\gamma_t(x)$ given the history $\bm{H}_t$ to its conditional density. The \textit{scoring functions} $g(x,t,k,\gamma_t(x)\mid \bm{H}_t)$, one for each event type $k$, emphasize an input event that is likely to occur and depreciates unsuspected events based on the past events $\bm{H}_t$. 
The subsequent functions $\phi(\,\cdot\,, k)$ serve as \textit{time scaling and decay function} for each event type $k$ and are non-negative. Finally, the output is passed through a \textit{sigmoid function} $\sigma$ to limit the range of the outcome. %The architecture of the entire model is illustrated in Fig.~\ref{fig:MNSTPP}.
% \begin{figure}[h!]
%     \centering
%     \includegraphics[width=.5\linewidth]{figures/TPPs.png}
%     \caption{\textcolor{red}{TBD: MNSTPP architecture}}
%     \label{fig:MNSTPP}
% \end{figure}

In the following Sec.~\ref{subsec:DGNN}, the DGNN that is used in the scoring function is introduced in detail, and afterwards, the scoring functions for different event types and the decay function is defined in Sec.~\ref{subsec:scoring_decay}.

\subsection{Dynamic Graph Neural Network}\label{subsec:DGNN}

In this model, a DGNN iteratively computes node and edge embeddings for all nodes and edges within the event stream. The item embeddings are calculated based on the event and its history. The embedding incorporates self-propagation, neighborhood propagation, exogenous drive, and attribute embedding modules to encompass all relevant information, formalizes as follows: 
\begin{align}\label{eq_node_embedding}
\bm{Z}^x_t = \sigma \left(\underbrace{\bm{M}_0\bm{h}_{loc}(x, t)    }_{\text{loc. embd. prop.}} +             \underbrace{\bm{M}_1 \bm{Z}^x_{\bar{t}}}_{\text{self-prop.}}  + 
\underbrace{\bm{m}_2 \cdot (t - \bar{t})}_{\text{exogenous drive}} +
\underbrace{\bm{M}_3 \cdot \bm{u}^x_t}_{\text{attribute prop.}} \right).
\end{align}

Depending on whether the item is a node $v$ or edge $\{u,v\}$, the embeddings vary in their local embedding propagation scheme using a different neighborhood definition. For a node, all neighboring nodes (connected by an edge), and for an edge $e=\{u,v\}$ the incident nodes $u,v$. The used item embeddings are further combined linearly and scaled by the attention factors $q$, which are computed as in GATv2 \cite{GATv2}:
\begin{align}
    \bm{h}_{loc}(x, t) = \sum_{w\in\mathcal{N}(x)} q^{\{x,w\}}_t \cdot \bm{Z}^w_{t}.
\end{align}

The self-propagation includes the evolution of the item embedding $\bm{Z}_t^x$ for an item $x$ at time $\bar{t}$, i.e., the previous time stamp of an event operating on the item $x$. The exogenous drive accounts for the time difference between the current event timestamp and the previous one, influencing an item at time $t$ and $\bar{t}$, respectively. 

The attribute embedding component incorporates the dynamic behavior of node/edge attributes via a Recurrent Neural Network (RNN). This RNN encodes historical attribute information in the embedding vector $\mathbf{u}^x_{\bar{t}}$, alongside the current attribute vector $\gamma_t(x)$, provided by a preprocessing module (e.g., CNN for images, text2vec\cite{text2vec} for text). 

The attribute update scheme for attribute $\gamma_t(x)$ of item $x$ at timestamp $t \in \mathcal{T}$, with learnable parameters $\bm{M}_4 \in \mathbb{R}^{d_h \times (d_{\gamma})}$ and $\mathbf{m}_5 \in \mathbb{R}^{d_h}$, is given by:

\begin{align}
    \bm{u}^x_t = \sigma\left(\bm{M}_4 \cdot \left[  \bm{u}^x_{\bar{t}} \| \gamma_t(x) \right] + \bm{m}_5\right).
\end{align}

In total, the proposed DGNN integrates the commonly used graph convolution and attention mechanisms in the local embedding propagation to consider the local structural neighborhood. In addition, the embeddings incorporate the local temporal information by using the previous embedding (i.e.) in the self-propagation and the exogenous drive. Furthermore, embedding involves attribute propagation and, thus, exploits all the information given in an event.
These embeddings are now used to define the scoring functions in the following subsection which determine the MNSTPP together with the decay function completely.
\subsection{Scoring Functions}\label{subsec:scoring_decay}

Given the embeddings of the historical and the current events, we now introduce the scoring functions that define the MNSTPP. Since we assume that the occurrence of events of different types have different temporal behaviors, we utilize one TPP, and thus one scoring function, per event type. In order to keep the notation short in the following, the scoring function $g(x,t,k,\gamma_t^x\mid \bm{H}_t)$ is denoted by $g_k(x,t)$.

\paragraph{Node Addition.} The scoring function for node addition events, such as $(v,t,k=0,\alpha_t(v))$, is learned in a linear layer that uses the node embedding from \ref{eq_node_embedding} at timestamp $t$:
\begin{align}
    g_0(v,t) = \bm{w}_0^\top \bm{Z}_t^v,
\end{align}
with learnable vector $\bm{w}_0\in\mathbb{R}^{d_h}$ and hidden dimension $d_h$. The node embedding includes the attribute embedding and thus involves the entire information available for this node.
The score is assumed to be high, when the event happens at time $t$ and low otherwise. In addition, the goal is to learn the scoring such that the score of impossible events as double additions or deletions is near zero.

\paragraph{Node Deletion.} For a node deletion event $(v,t,k=1,\alpha_t(v))$, the current node embedding $\bm{Z}_t^v$ contains the attribute and neighborhood information, as well as the historical information of the node. So the score
\begin{align}
    g_1(v,t) = \bm{w}_1^\top \bm{Z}_t^v
\end{align}
with learnable vector $\bm{w}_1\in\mathbb{R}^{d_h}$ considers the local structural and temporal information of the node. 

\paragraph{Edge Addition.} The addition of an edge $\{u,v\}$ is assumed to be dependent on the possible incident nodes $u,v$. Hence, the scoring function for edge addition events like ${(\{u,v\},t,k=2,\beta_t(\{u,v\}))}$ considers the corresponding node and edge embeddings at time $t$ and is determined by
\begin{align}
    g_2(\{u,v\},t) &= \bm{W}_2^\top\cdot \text{ReLU}\Bigl(\hat{\bm{W}}_2^\top\cdot \hat{\bm{Z}}_t^{N(\{u,v\})}\Bigr), \\
    \text{with\quad}\hat{\bm{Z}}_t^{\set{N}(\{u,v\})} &= [\bm{Z}_t^u\| \bm{Z}_{t}^v \| \bm{Z}_{t}^{\{u,v\}}]\nonumber
\end{align}
and $\bm{W}_2\in\mathbb{R}^{d_h\times 1}$ and $\hat{\bm{W}}_2\in\mathbb{R}^{(3 d_h)\times d_h}$. Instead of one forward layer, an MLP with two layers is used to allow for more detailed learning.

\paragraph{Edge Deletion.} Analoguously to the edge addition, the scoring function for an edge deletion event ${(\{u,v\},t,k=3,\beta_t(\{u,v\}))}$ utilizes the embeddings of the incident nodes at time $t$, here together with the \textit{previous} edge embedding $\bm{Z}_{\bar{t}}^{u,v}$:
\begin{align}
    g_3(\{u,v\},t) &= \bm{W}_3^\top\cdot \text{ReLU}\Bigl(\hat{\bm{W}}_3^\top\cdot \hat{\bm{Z}}_t^{\set{N}(\{u,v\})}\Bigr), \\
    \text{with\quad}\hat{\bm{Z}}_t^{\set{N}(\{u,v\})} &= [\bm{Z}_t^u\| \bm{Z}_{t}^v \| \bm{Z}_{\bar{t}}^{\{u,v\}}]\nonumber
\end{align}
and $\bm{W}_3\in\mathbb{R}^{d_h\times 1}$ and $\hat{\bm{W}}_3\in\mathbb{R}^{(3 d_h)\times d_h}$.

\paragraph{Node Attribute Change.} The scoring function of an event $(v,t,k=4,\alpha_t(v))$ that changes the previous attribute $\alpha_{\bar{t}}(v)$ of node $v$ to $\alpha_t(v)$ incorporates the corresponding attribute and node embeddings to explicitely take the attribute change into account:
\begin{align}
     g_4(v, t) =  \underset{attribute\ score}{\underbrace{\bm{W}_4^\top\cdot \text{ReLU}\Bigl(\hat{\bm{W}}_4^\top\cdot \bm{u}_t^v\Bigr)}} + 
        \underset{embedding\ score}{\underbrace{\bar{\bm{W}}_4^\top\cdot \text{ReLU}\Bigl(\tilde{\bm{W}}_4^\top\cdot \bm{Z}_t^v\Bigr)}}
\end{align}
with $\bm{W}_4,\bar{\bm{W}}_4\in\mathbb{R}^{d_h\times 1}$ and $\hat{\bm{W}}_4, \tilde{\bm{W}}_4\in\mathbb{R}^{ d_h\times d_h}$. The two-layered MLPs are used to avoid overflow of the summands.

\paragraph{Edge Attribute Change.} Bringing together the ideas of the scores for edge additions/deletions and node attribute changes, the scoring function for an edge attribute change $(\{u,v\},t,k=5,\beta_t(\{u,v\}))$ combines an attribute, an embedding and a neighborhood score as follows:
\begin{align}
        g_5(\{u,v\}, t)  &= \underset{attribute\ score}{\underbrace{\bm{W}_5^\top\cdot \text{ReLU}\Bigl(\hat{\bm{W}}_5^\top\cdot \bm{u}_t^{\{u,v\}}\Bigr)}} + 
        \underset{embedding\ score}{\underbrace{\bar{\bm{W}}_5^\top\cdot \text{ReLU}\Bigl(\tilde{\bm{W}}_5^\top\cdot \bm{Z}_t^{\{u,v\}}\Bigr)}} \nonumber\\
        &\quad + \underset{neighborhood\ score}{\underbrace{\bar{\bar{\bm{W}}}_5^\top\cdot \text{ReLU}\Bigl(\hat{\hat{\bm{W}}}_5^\top\cdot \hat{\bm{Z}}_t^{\set{N}(\{u,v\})}\Bigr)}}\\
        \text{with }\hat{\bm{Z}}_t^{\set{N}(\{u,v\})} &= [\bm{Z}_t^u\| \bm{Z}_{t}^v \| \bm{Z}_{t}^{\{u,v\}}]\nonumber
\end{align}
and weights $\bm{W}_5, \bar{\bm{W}}_5, \bar{\bar{\bm{W}}}_5\in\mathbb{R}^{d_h\times 1}$, $\hat{\bm{W}}_5^, \tilde{\bm{W}}_5\in\mathbb{R}^{ d_h\times d_h}$ and $\hat{\hat{\bm{W}}}_5\in\mathbb{R}^{3 d_h\times d_h}$.

The defined scoring functions can potentially have any real value. One way to limit them is to first forward the scores to a decay function that restricts the scoring to be positive, and second to squash the outcome of the entire model into the unit interval by a sigmoid activation as already mentioned in eq.~\eqref{eq:MNSTPP}.

\subsection{Decay Function}
The last module of the MNSTPP consists of a temporal decay function that also serves as a time scaling procedure. Assuming that 1) information gets obsolete after time and 2) the event frequencies may vary between the event types, we use one decay function for each event type with a learnable time-scaling parameter that is, analoguously to \cite{inproc_trivedi_2019}, similar to the softplus function:
\begin{align}
    \phi_k(x)=\psi_k \text{log}(1 + \text{exp}(x/\psi_k))
\end{align}
with timescale parameter $\psi_k > 0$ for each event type $k$.

Coming back to the entire chain of functions in eq.~\eqref{eq:MNSTPP}, the MNSTPP model consists of six Marked Neural Spatio-Temporal Point Processes with corresponding intensity functions $\lambda_k$ for short, that incorporate the local structural and temporal information of the events. As a consequence, they represent processes that describe the node and edge events occurring in a graph stream and characterize the conditional density for events to happen in a short time period. In the following section, we describe the training and prediction procedures utilizing the derived intensity functions.

\subsection{Embedding Initialization and Update}
For the start graph as well as added nodes and edges, the model needs initial embeddings for further calculations. First, the start embeddings of all nodes and edges $x\in\set{V}_0\cup\set{E}_0$ in the start graph are set to the corresponding attribute embedding $\bm{Z}_0^x:=\tilde{\bm{u}}_0^x=\sigma(\bm{M}_x\gamma_0^x)$ with seperate weights $\bm{M}_N$ for nodes and $\bm{M}_E$ for edges. If the nodes or edges have no attributes or they all have the same attribute, a small noise between $[-1,1]$ is added. Then, they are updated with a graph convolution $\bm{Z}_0^x = \sigma \bigl(\bm{M}_0\bm{h}_{loc}(x,0) + \bm{M}_3\cdot\tilde{\bm{u}}_0^x\bigr)$ with weights from Eq.~\eqref{eq_node_embedding}. In contrast, if a node $v$ gets added at time $t$ in the event stream, the embeddings are initialized with $\tilde{\bm{u}}_t^v$. For edge additions, the embeddings are calculated analoguously to the edge embeddings in the start graph.

If edges are added or deleted in the event stream, the neighborhood of the incident nodes change immediately. Therefore, the embeddings of the incident nodes are updated on the fly by recalculating the local embedding propagation and overwriting the embedding respectively.

\section{Training and Prediction}
\subsection{Training}
The training is performed similarly to the procedure described in \cite{inproc_trivedi_2019}. Here, we have to extend the loss function to include the processing of continuous marks in the intensities. The learning is then conducted by minimizing the negative log-likelihood of observed events $\set{O}$ together with the averaged intensities of unobserved events $\tilde{\mathcal{O}}_t$ until time $t$ (so-called\textit{ negative samples}, see Sec.~\ref{sec:negative_sampling}):
\begin{align}\label{training}
    \mathcal{L} &:= - \sum_{o\in \mathcal{O}} \text{log} \left( \lambda(o) \right) + \underset{\text{\scriptsize\textcolor{gray}{survival probability}}}{\underbrace{\frac{\rho}{|\tilde{\mathcal{O}}_t|}\sum\limits_{\tilde{o}\in\tilde{\mathcal{O}}_t} \lambda(\tilde{o})}}.
\end{align}
Here, the penalty $\rho\geq 0$ regularizes the proportion from the influence of negative samples to the observed events.

\subsection{Negative Sampling}\label{sec:negative_sampling}
Per event, we sample a \textit{logical} and a \textit{non-logical negative sample}. Logical negative samples comprise events that may happen but have not been observed (yet). E.g., a node of the graph gets deleted in a negative event, even though it further exists, or an edge attribute is changed in a negative event, even though the attribute stays the same. In contrast, non-logical samples include events that are impossible, such as adding a node that already exists, deleting a non-existent edge or changing an attribute of a node that does not exist. We create the \textbf{logical negative samples} in the following way:
\begin{itemize}
    \item \underline{\textbf{Node/edge addition:}}  For a negative event $(x,t,k=0/2,\gamma_t^x)$ to an observed addition event $o$, a node $x$ is set to the current maximal number id $+\,1$ or an edge $x$ is randomly chosen from the current non-existent edges. The timestamp $t$ is randomly chosen within a small interval around the timestamp of $o$ and the attribute $\gamma_t^x$ is set to the attribute from a randomly chosen existent node or edge. 
    \item \underline{\textbf{Node/edge deletion:}} A negative deletion event is created by choosing a random existing node or edge, take the corresponding attribute and select the timestamp as in the addition case.
    \item \underline{\textbf{Node/edge attribute change:}} To generate a negative sample for an attribute change, an existent node or edge is chosen randomly, the timestamp is created as in the addition case, and the new attribute is set set to a vector close to the attribute from the observed event $o$. For this purpose, a vector is sampled from the hypersphere with center in the attribute from $o$ and a certain radius.
\end{itemize}
For the \textbf{non-logical negative samples}, the procedure is similar but we ensure that we choose existent nodes or edge to be added and non-existent nodes or edges to be deleted, as well as non-existent nodes or edges whose attributes are changed in the negative samples.

\subsection{Batch-wise Processing}\label{sec:batch_processing}
As in \cite{chen2023neutronstream}, the input graph is batched respecting dependent events. The events are grouped in different batches if they occur close to each other in the graph, i.e., if nodes are neighbors or edges have a common node, the events are processed in different batches. This way, a set of events can be processed in parallel in one batch without causing access and overwriting problems in the implementation.
\subsection{Prediction}
Let $\bm{m}=\{x,k,\gamma\}\in (\set{V}\cup\set{E})\times\{0,\ldots,5\}\times\mathbb{R}^d=:\mathcal{M}$ be the marks with item $x$, event type $k$ and attribute $\gamma\in\mathbb{R}^d$, and $t \geq 0$ a time stamp.
Then, the conditional probability density function can be chosen as 
\begin{align}\label{equation_density_fnk_structural_change}
f(t,\bm{m}\mid \bm{H}_{\bar{t}}) & = \frac{\lambda(t, \bm{m}\mid \bm{H}_{\bar{t}})}{1- F(t,\bm{m}\mid \bm{H}_{\bar{t}})} \; \text{ interpretation of Eq.~\eqref{eq_MTPP} and \cite{STPP_book}}\nonumber\\
&= \lambda(t, \bm{m}\mid \bm{H}_{\bar{t}}) \cdot \text{exp}\left(-\int_{\mathcal{M}}\int_{[\bar{t},t]} \lambda( \tau, \bm{m}\mid \bm{H}_{\bar{t}})\, \text{d}\tau\text{d}\bm{m}\right) \; \text{log rule, see \cite{TPP_lecture}}\nonumber\\
&= \lambda(t, \{x,k,\gamma\}\mid \bm{H}_{\bar{t}}) \cdot \text{exp}\left(-\sum_{\xi\in\set{V}\cup\set{E}}\sum_{\kappa\in[5]}\int_{\mathbb{R}^d}\int_{[\bar{t},t]} \lambda( \tau, \xi, \kappa, \delta\mid \bm{H}_{\bar{t}}) \,\text{d}\tau\text{d}\delta\right)\nonumber
\end{align}
%considering an appropriate intensity function $\lambda_k$. It provides the occurrence probability of an event of type $k$ in the time interval $[\bar{t}, t]$ given that the underlying process only includes temporal dependencies. 

%\textcolor{red}{Naja hier fehlt ja noch einiges. Man kann admit ja nicht nur die structural changes predicten, sondern jenachdem was man festhält kann man ja predicten: Was für ne dynamic als nächstes passiert auf einem bestimmten knoten/kante (indem man das maximum über alle k nimmt), und sogar auf welchem knoten oder kante (indem man das maximum nimmt über alle knoten/kanten)}
%Note that Eq.~\eqref{equation_density_fnk_structural_change} is not appropriate to predict the node appearance for nodes that have never existed. This case has to be investigated further. 
%While using active or inactive nodes or edges does not cause any problems, adding new nodes that never existed in the graph event history Eq.~\eqref{equation_density_fnk_structural_change} seems inappropriate, which we want to investigate further.
The conditional density function $f(t,x,k,\gamma)$ determines a joint distribution over the time, the nodes/edges, the event types and the attributes.
To predict the next future timestamp $t^*$ when an event of type $k$ at an item $x$ occurs, we consider the expectation given the history $\bm{H}_t$. For this purpose, the conditional density function is integrated over the future starting at the previous timestamp $\bar{t}$ and the attribute space $\mathbb{R}^d$ given by
\begin{align}
    t^* = \mathbb{E}[t \mid \bm{H}_t, k, x] &=   \int\limits_{[\bar{t},\infty]}\int\limits_{\mathbb{R}^d} \tau \cdot f(\tau, x, k, \bm{\delta})  \text{d}\bm{\delta}\text{d}\tau\nonumber\\
    &=  \int\limits_{[\bar{t},\infty]} \tau\cdot \int\limits_{\mathbb{R}^d}  \lambda(\tau, x, k, \bm{\delta}) \cdot \text{exp}\left(-\int\limits_{\mathcal{B(\bm{\delta})}}\,\int\limits_{[\tau,\tau+\Delta\tau]} \lambda( \tau_1, x, k, \bm{\delta}_1)\,\text{d}\tau_1\,\text{d}\bm{\delta}_1 \right) \text{d}\bm{\delta}\text{d}\tau,\nonumber
\end{align}
where $\mathcal{B}(\bm{\delta})\subset\mathbb{R}^d$ is an open ball with center $\bm{\delta}$ and radius $\varepsilon>0$, and $[\tau,\tau+\Delta\tau]\subset\mathbb{R}$ is a time interval. 
We have to integrate over the attribute space, since the attribute of an item is dependent on the time. Further, given the item $x$, the type pf event $k$ and the history up to time $t$, the item attribute can be predicted analogously with the expectation value
\begin{align}
    \bm{\gamma}^* = \mathbb{E}[\bm{\gamma} \mid \bm{H}_t, k, x] &=  \int\limits_{\mathbb{R}^d} \int\limits_{[\bar{t},\infty]}\bm{\delta} \cdot f(\tau, x, k, \bm{\delta})  \text{d}\tau\text{d}\bm{\delta}\nonumber\\
    &=  \int\limits_{\mathbb{R}^d} \bm{\delta} \cdot \int\limits_{[\bar{t},\infty]} \lambda(\tau, x, k, \bm{\delta}) \cdot \text{exp}\left(-\int\limits_{\mathcal{B(\bm{\delta})}}\,\int\limits_{[\tau,\tau+\Delta\tau]} \lambda( \tau_1, x, k, \bm{\delta}_1)\,\text{d}\tau_1\,\text{d}\bm{\delta}_1 \right) \text{d}\tau\text{d}\bm{\delta}.\nonumber
\end{align}

\begin{align}
    k^*|_x &= \underset{\kappa\in\{0,\ldots5\}}{\text{argmax}}\int\limits_{[\bar{t},\infty]} \int\limits_{\mathbb{R}^d} f(\tau, x, \kappa, \bm{\delta}) \text{d}\bm{\delta}\text{d}\tau 
    %&= \int\limits_{[t,\infty]} \left(-\sum_{\xi\in\set{V}\cup\set{E}}\sum_{\kappa\in[5]}\int_{\mathbb{R}^d}\int_{[\bar{t},t]} \bm{\delta}\cdot \lambda( \tau, \xi, \kappa, \bm{\delta}\mid \bm{H}_{\bar{t}}) \,\text{d}\tau\text{d}\bm{\delta}\right)
\end{align}
$|_x$ and $|_k$ indicate that we have $x$ or $k$ given. Further, the sets $\{0,\ldots,5\}$ and $\set{V}\cup\set{E}$ considered here are dependent on the choice of $x$ or $k$, respectively, since some event types only occur on nodes and some only on edges, and so they determine each other.
\begin{align}
    x^*|_k &= \underset{\xi\in\set{V}\cup\set{E}}{\text{argmax}}\int\limits_{[\bar{t},\infty]} \int\limits_{\mathbb{R}^d} f(\tau, \xi, k, \bm{\delta}) \text{d}\bm{\delta }\text{d}\tau.
    %&= \int\limits_{[t,\infty]} \left(-\sum_{\xi\in\set{V}\cup\set{E}}\sum_{\kappa\in[5]}\int_{\mathbb{R}^d}\int_{[\bar{t},t]} \bm{\delta}\cdot \lambda( \tau, \xi, \kappa, \bm{\delta}\mid \bm{H}_{\bar{t}}) \,\text{d}\tau\text{d}\bm{\delta}\right)
\end{align}

%\textcolor{blue}{$f()$ is a joint distribution over time and attribute space, where the distributions have different forms dependent on the event type and node/edge. Since we assume to have similar event behaviors on the nodes and edges, the TPP is only a process in time with iid marks that only depent on the timestamp.}

The double integrals are approximated with Monte Carlo sampling as proposed in \cite{mishra2023solving}.

\paragraph{Remark.} Note that if we want to predict an item $x$ for an add node event ($k=0$), the node is always set to the next available node id, so this prediction is unnecessary. However, predicting \textit{when} or with \textit{which attribute} a node is added, is appropriate.

%\section{Experiments}
%\input{Experiments/datasets}
%\input{Experiments/baselines}
%\input{Experiments/experimental_setup}
%\input{Experiments/evaluation}

%\input{other_sections/discussion}

\section{Conclusion and Future Work}

The proposed model enables learning on graph streams, including arbitrary types of dynamics. The utilized Dynamic GNN processes the historical structural and temporal information and provides hidden representations of the nodes and edges. The subsequent different TPPs model the temporal behavior of the graph stream based on the hidden graph representation, and new occurring events can be efficiently integrated into the model. 

However, our model is a work in progress and thus provides opportunities for further development. The model's reliability, explainability, and generalization are briefly discussed in the following.

\textbf{Reliability.} The proposed work still needs to be evaluated. Hence, extensive experiments must be conducted to prove the applicability and reliability of the model.

\textbf{Explainability.} 
GNNs are called explainable if the model explains the predicted result or reasoning that can be inferred based on the model architecture. In the future, the goal is to make our model more explainable to simplify the application of the model for real-world problems. %For this purpose, including more components in the architecture that improve explanations for the model output considering the input graphs is necessary. Initially, the attention mechanisms (cf.~§\ref{section_attentions}) allow for a better understanding of the model's behavior.

% \textbf{Definite Deletions vs.~Inactivity.}
% Furthermore, the activity function from Section \ref{section_node_edge_activities} will be extended to consider definite deletions, activities, and inactivities of nodes/edges. Thereby, nodes and edges lose the possibility to occur again in the graph.

\textbf{More complex graph types.}
The structural graph properties of the dynamic graphs in this paper are yet elementary, so the MNSTPP can be extended to more complex dynamical graph structures in the future, considering, e.g., the graph types described in \cite{jour_thomas_2021}.

%\textbf{Self-Propagation.} Is possible.

\section*{Acknowledgement}
AMO, SBW, YN and JT are funded by the Ministry of Education and Research Germany (BMBF), under the funding code 01IS20047A, according to the 'Policy for funding female junior researchers in Artificial Intelligence'. \\
Further, the authors would like to thank Prof.~Dr.~Felix Lindner and Jan Schneegans for the fruitful discussions and helpful feedback on our work. 

\section*{Contribution}\label{apx:contribution}

\begin{multicols}{2}
\begin{itemize}
    \item Conceptualization: SBW, AMO
    \item Methodology: SBW, AMO, RN
    \item Resources: AMO
    \item Writing (Original Draft): SBW, AMO
    \item Writing (Review \& Editing): SBW, AMO, RN, YN
    \item Supervision: RN, JT
    \item Project administration: SBW, AMO
    \item Funding acquisition: JT
\end{itemize}
\end{multicols}

Alice Moallemy-Oureh is mainly responsible for incorporating attribute dynamics, while Silvia Beddar-Wiesing is mainly responsible for modeling the structure dynamics.

\newpage
\bibliographystyle{abbrv}
\small{
\bibliography{main.bib}

%\appendix
%\input{appendix/appendix}
}

%%%%%%%%%%%%%%%%%%%%%%%%%%%%%%%%%%%%%%%%%%%%%%%%%%%%%%%%%%%%

\end{document}